\documentclass[10pt,twocolumn,letterpaper]{article}

\usepackage{iccv}
\usepackage{times}
\usepackage{epsfig}
\usepackage{graphicx}
\usepackage{amsmath}
\usepackage{amssymb}

\usepackage{verbatim}
\usepackage{multirow}

\usepackage{listings}
\usepackage{url}
\usepackage{tabularx}
\usepackage{booktabs}
\usepackage{amsmath,amssymb,amsfonts}
\usepackage[accsupp]{axessibility} 

\usepackage[pagebackref=true,breaklinks=true,colorlinks,bookmarks=false]{hyperref}

\newcommand{\matr}[1]{\mathbf{#1}}     

\makeatletter
\DeclareRobustCommand\onedot{\futurelet\@let@token\@onedot}
\def\@onedot{\ifx\@let@token.\else.\null\fi\xspace}

\def\eg{\emph{e.g}\onedot} 
\def\ie{\emph{i.e}\onedot}

\def\etal{\emph{et al}\onedot}

\def\wrt{w.r.t\onedot}

\def\etal{\emph{et al}\onedot}

\makeatother
\def\T{^{\mathsf T}}
\newcommand{\Real}{\mathbb{R}}
\DeclareMathOperator*{\E}{\mathbb{E}}
\DeclareMathOperator*{\onehot}{one\_hot}
\DeclareMathOperator*{\softmax}{softmax}
\DeclareMathOperator*{\topk}{top-{\it k}}

\usepackage{siunitx, cleveref}
\crefname{figure}{Fig.}{Figs.}
\renewcommand{\paragraph}[1]{\vspace{0.5em}\noindent\textbf{#1}}
\usepackage[breaklinks=true,bookmarks=false]{hyperref}
\iccvfinalcopy 

\ificcvfinal\fi

\begin{document}

\title{Generalized Differentiable RANSAC}

\author{Tong Wei$^1$, Yash Patel$^1$, Alexander Shekhovtsov$^1$, Ji{\v{r}}{\'\i} Matas$^1$, and Daniel Barath$^2$ \\
$^1$ Visual Recognition Group, FEE, Czech Technical University in Prague \\$^2$ Computer Vision and Geometry Group, ETH Zurich \\
{\tt\small \{weitong, patelyas, shekhole, matas\}@fel.cvut.cz, danielbela.barath@inf.ethz.ch}}

\maketitle
\ificcvfinal\thispagestyle{empty}\fi


\begin{abstract}
We propose $\nabla$-RANSAC, a generalized differentiable RANSAC that allows learning the entire randomized robust estimation pipeline. 
The proposed approach enables the use of relaxation techniques for estimating the gradients in the sampling distribution, which are then propagated through a differentiable solver.
The trainable quality function marginalizes over the scores from all the models estimated within $\nabla$-RANSAC to guide the network learning accurate and useful inlier probabilities or to train feature detection and matching networks.
Our method directly maximizes the probability of drawing a good hypothesis, allowing us to learn better sampling distributions.
We test $\nabla$-RANSAC on various real-world scenarios on fundamental and essential matrix estimation, and 3D point cloud registration, outdoors and indoors, with handcrafted and learning-based features.
It is superior to the state-of-the-art in terms of accuracy
while running at a similar speed to its less accurate alternatives.
The code and trained models are available at \url{https://github.com/weitong8591/differentiable\_ransac}.

\end{abstract}

\section{Introduction}
\label{sec:intro}

Robust estimation is a fundamental component in vision pipelines, including relative pose estimation~\cite{chum2004epipolar}, 
wide baseline matching~\cite{pritchett1998wide,matas2004robust,mishkin2015mods}, 
multi-model fitting~\cite{isack2012energy,pham2014interacting}, 
image-based localization~\cite{Brachmann_2017_CVPR}, motion 
segmentation~\cite{torr1993outlier}, 
and pose graph initialization of Structure-from-Motion (SfM) algorithms~\cite{schoenberger2016sfm, schoenberger2016mvs}.
While several robust estimators have been proposed throughout the years~\cite{holland1977robust,illingworth1988survey,li2009consensus,xiao2016hypergraph}, randomized hypothesize-and-verify approaches, like RANSAC~\cite{RANSAC} and its recent variants~\cite{barath2018graph,barath2019magsac,barath2020magsac++,ivashechkin2021vsac}, have become the most widely used methods due to their robustness, simplicity, and efficiency. 
RANSAC repeatedly selects random minimal subsets of the input data sufficient to fit a model hypothesis, \eg, a 3D plane to three points or a fundamental matrix to seven point correspondences.
The model score is then computed as the cardinality of the inlier set, formed by the points consistent with the model hypothesis, \ie, having residuals smaller than a threshold. 
The so-far-the-best model is updated if a new model is found with higher quality. RANSAC terminates when the probability of finding a better hypothesis falls below a threshold.
Finally, the model with the highest quality, polished, \eg, by least-squares fitting on all inliers, is returned.

\begin{figure*}[t]
	\centering
    \includegraphics[width = 2.1\columnwidth, trim=0mm 0mm 0mm 0mm, clip]{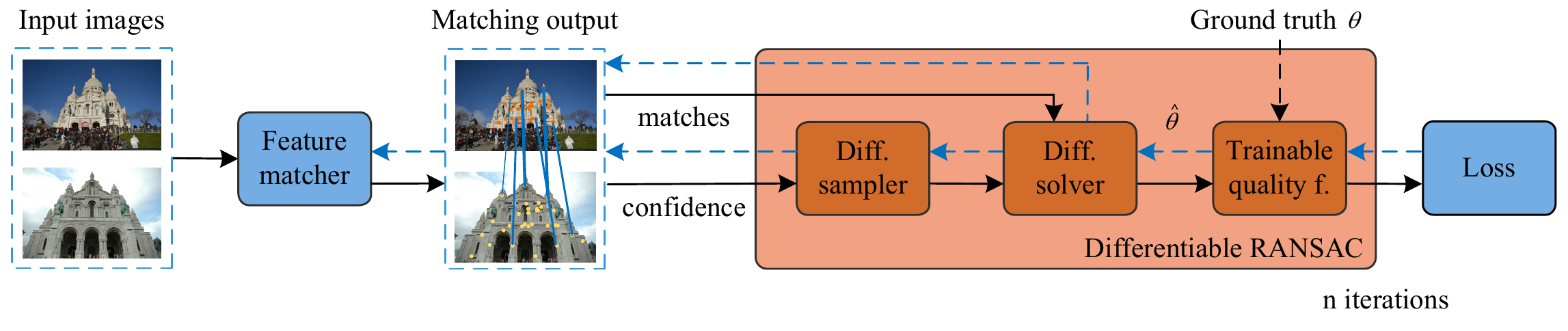}
    \vspace{-1em}
    \caption{\textbf{Pipeline of training $\nabla$-RANSAC (forward and backward).}
    Given an image pair, we can either use a hand-crafted feature matcher and feed the tentative correspondences into the consensus learning network from \cite{clnet2021}, or use learning-based matching method that outputs matches and their confidence.
    The predictions input to the $\nabla$-RANSAC module for robust estimation.
    In each iteration, the differentiable and randomized Gumbel sampler (Section~\ref{sec:sampler}) selects a minimal sample of $m$ correspondences. 
    Model $\hat{\theta}$ is estimated by differentiable solvers (Section~\ref{sec:diff_solver}) and its loss is calculated based on trainable quality functions (Section~\ref{sec:quality}) with the ground truth. }
   \label{fig:flowchart}
   \vspace{-1em}
\end{figure*}

Numerous improvements have been made to the original RANSAC algorithm, including refinement of hypotheses through local optimization~\cite{chum2003locally,barath2018graph}, better scoring~\cite{torr2000mlesac,barath2019magsac,barath2020magsac++}, detection of degenerate cases~\cite{chum2005two}, and speed-ups through techniques such as weighted random sampling of hypotheses~\cite{chum2005matching} or preemptive hypothesis verification strategies~\cite{chum2002randomized,chum2008optimal}.
See~\cite{imc2022} for a recent survey and benchmark.
To \cite{imc2022}, the most accurate method for relative pose estimation is MAGSAC++~\cite{barath2020magsac++} with PROSAC sampling~\cite{chum2005matching} and DEGENSAC-based degeneracy~\cite{chum2005two} testing. 

In recent years, neural networks (NNs) have been employed to estimate tentative matches, including their coordinates and confidences~\cite{cne2018,acne2020,oanet2019,clnet2021,Dai_2022_CVPR,Sun_2021_CVPR}. 
These predicted confidences could be used for pre-filtering matches or weighted random sampling in RANSAC. 
However, learning these NNs endowed with RANSAC, particularly for optimizing the desired evaluation metric, such as pose error, remains a challenging problem. 
One of the main challenges is that minimal solvers are often complex and not readily differentiable. Additionally, learning the sampling distribution for optimal RANSAC performance is challenging, both in terms of formalizing the problem and estimating gradients for the sampling probabilities. 
Prior work in this direction~\cite{Brachmann_2017_CVPR,brachmann2019neural} will be discussed in detail (Section~\ref{sec:related}).

In this paper, we make contributions to the learnable robust estimation family and propose a new differentiable RANSAC, $\nabla$-RANSAC, with all the components differentiable. 
The main contributions are as follows:
\begin{itemize}
    \setlength\itemsep{-0.5mm}
    \item We investigate all the components of the RANSAC pipeline and propose a new differentiable alternative that allows learning inlier probabilities while directly optimizing test-time evaluation metrics, \eg, pose error, as a \textit{new} learning objective.  
    \item We propose a new random sampling approach based on a re-parametrization strategy, \ie Gumbel Softmax sampler, that allows gradient propagation through the entire randomized procedure.
    \item To demonstrate its potential to unlock the end-to-end training of geometric pipelines, $\nabla$-RANSAC is incorporated into an end-to-end feature matcher, LoFTR \cite{Sun_2021_CVPR}, to improve the predicted matches and confidence.
    \item Technically, we implement and include a  differentiable version of the widely used minimal solvers, \eg, five and seven-point algorithms~\cite{nister2004efficient}, and standard Kabsch algorithm~\cite{kabsch1976solution} for rigid transformation during training.
\end{itemize}
$\nabla$-RANSAC has significant implications for learning-based vision systems, enabling training such pipelines that were previously difficult or impossible to train. 
\section{Related Work}
\label{sec:related}
\paragraph{Robust Estimation with NNs.}
Context normalization networks (PointCN)~\cite{cne2018} is one of the first papers on the topic, using a PointNet-based ~\cite{qi2017pointnet} structure with batch normalization~\cite{batchnorm2015} as a context mechanism to predict inlier probabilities.
Attentive context normalization networks~\cite{acne2020} improve upon \cite{cne2018} by using a special architectural block.
Deep Fundamental matrix estimation~\cite{dfe2018} iteratively estimates the model using weighted least squares (LS) with weights suppressing the effect of outliers, re-estimated in each iteration.
\cite{oanet2019,Dai_2022_CVPR} employ NNs to filter outliers, while CLNet~\cite{clnet2021} estimates the weights to be used in progressive filtering and weighted LS.
Although many of these methods use weighted LS for model estimation due to its easy gradient propagation, it has been shown that applying RANSAC on correspondences filtered by CLNet or similar techniques improves results~\cite{cvpr2020ransactutorial}. 
In other words, RANSAC is still necessary to robustly verify ambiguous hypotheses.

\paragraph{Sampling Distribution.}
In the worst case, the probability of drawing a good hypothesis at random in RANSAC decreases exponentially with the minimal sample size.  
Consequently, RANSAC might take excessive time or return an inaccurate solution if stopped early.
It was observed that using a weighted random sampling~\cite{chum2005matching,brachmann2019neural,barath2020magsac++}, which is more likely to draw inlier points, often significantly improves the performance. 
This sampling guidance can either come from the feature matching procedure, \eg, as the SNN ratio~\cite{lowe1999object} or can be learned from ground truth inliers~\cite{brachmann2019neural}.
PROSAC~\cite{chum2005matching} exploits estimated inlier probabilities to sample the most promising hypotheses first.
P-NAPSAC~\cite{barath2020magsac++} progressively increases the radius of the selection hypersphere according to every unsuccessful iteration,  blending into global sampling.
Such samplers rely on the given sampling distribution, which might be improved using NNs.

\paragraph{Differentiable RANSAC.}
Currently, only one method learns the sampling distribution specifically for RANSAC, the Neural-Guided RANSAC~\cite{brachmann2019neural}. 
NG-RANSAC maximizes the expected quality of the model found by RANSAC end-to-end by applying {\sc reinforce}~\cite{Williams1992} gradient estimator. 
This unbiased estimator requires neither the solver nor the loss function to be differentiable. 
However, to improve the geometric features of individual tentative correspondences, backpropagation through the solver becomes necessary.
It can be achieved by numerically differentiating the solver, using the finite difference method proposed in DSAC~\cite{Brachmann_2017_CVPR} for the 4-point P\textit{n}P solver.
NG-DSAC~\cite{brachmann2019neural} combines these two techniques to jointly learn the sampling distribution and refine the coordinates.

Our work differs in several ways. 
First, we implement differentiable solvers for common minimal problems, enabling us to learn geometric features. 
In addition, this enables the use of relaxation techniques such as Gumbel-Softmax (GS)~\cite{jang2016categorical} to estimate the gradient in the sampling distribution instead of using {\sc reinforce}~\cite{Williams1992}. 
Towards this end, we propose a simple GS-like relaxation for drawing a minimal sample of $k$ points without replacement, which performs well in practice in our experiments. 
However, we remark that other (more complex) estimators, such as NeuralSort~\cite{NeuralSort}, can also be applied, enabled by the differentiable solvers.
Another key difference is a new objective function that maximizes the quality of an average sampled model instead of using the best one in the pool (NG-RANSAC). 
Our objective directly maximizes the probability of drawing a good hypothesis, allowing $\nabla$-RANSAC to better learn the sampling distribution. 
Finally, unlike previous work~\cite{brachmann2019neural}, we apply our method to jointly learn the sampling distribution and feature matches for epipolar geometry estimation.

\section{Generalized Differentiable RANSAC}
\label{sec:differentiable_RANSAC}

In this section, we discuss the algorithmic components of the proposed framework, visualized in \cref{fig:flowchart}. The key component is the differentiable $\nabla$-RANSAC block.

We assume that the input to $\nabla$-RANSAC is a set of tentative correspondences, possibly equipped with extra information from the detector and matcher, \eg, feature orientation, scale, affine shape, jointly referred to as {\em geometric features} $\Phi \in\Real^{N\times D}$ and their confidences represented by scores $s \in \Real^N$, where $N$ is the number of matches and $D$ is the number of geometric features per correspondence.
From the input image pairs, we adopt leaning-based feature matching methods, such as LoFTR, consensus learning CLNet architecture~\cite{clnet2021} to generate tentative point correspondences and confidence scores $s$.
These scores are originally used for iterative pruning of matches~\cite{clnet2021}, and we repurpose them for weighted sampling in RANSAC.

At test time, $\nabla$-RANSAC draws minimal samples of $k$ correspondences using weighted random sampling with probabilities $p = \softmax(s)$. The correspondences are drawn from the categorical distribution ${\rm Cat}(p)$ one-by-one without replacement until a minimal sample $(i_1,\dots i_k)$ of $k$ correspondences is formed $h = (\Phi_{i_1},\dots, \Phi_{i_k})$. Namely, the probability to draw a minimal sample $(i_1, \dots i_k)$ follows the Plackett-Luce (PL) model~\cite{Luce2012}:
\begin{align}
    \textstyle p(i_1) \frac{p(i_2)}{1 - p(i_1)} \dots \frac{p(i_k)}{1 - \sum_{l<k}p(i_l)}.
\end{align}
Then a {\em solver} returns solutions for geometric model $\hat \theta$, fitting precisely the minimal sample. The best model is selected, \eg, based on MAGSAC score~\cite{barath2020magsac++}, and its loss, \eg, the relative pose error, is evaluated \wrt the ground truth. 

 \begin{figure}[t]
	\centering
    \includegraphics[width = 0.99\columnwidth, trim=0mm 0mm 0mm 0mm, clip]{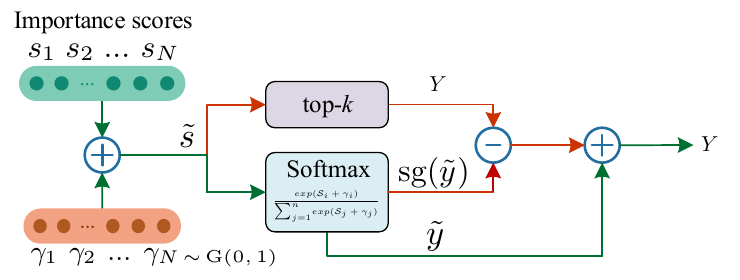}
    \vspace{0.5em}
    \caption{Overview of \textbf{Gumbel Softmax Sampler} 
 used in $\nabla$-RANSAC. The input to the sampler is the importance scores $s_1, s_2, ..., s_N$. The process starts by i.i.d random sampling $\gamma_1, \gamma_2, ..., \gamma_N$ from the standard ${\rm Gumbel}(0,1)$ distribution. Then a minimal sample is drawn by selecting indices of top $k$ noisy scores.
  Since $\topk$ is non-differentiable, the straight through trick is used and the Softmax output is added and subtracted with stop-gradient ($\rm sg$) to allow backward flow of the gradients.
 The green arrows show the connections through which gradient flow is possible. 
 Red arrows show connections without any gradients.
 }
    \label{fig:gumbel_sampler}
    \vspace{-0.5em}
\end{figure}

At training time, we are interested in learning the NN that produces geometric features and importance scores. 
We argue that learning importance scores for the best model in the entire RANSAC is impractical. If RANSAC runs long enough, it likely finds a good model independently of the confidence scores. The gradient in the scores is vanishing in the expectation and has high variance. Note that Brachmann {\em et al.}~\cite{brachmann2019neural}, while considering training the complete RANSAC algorithm, actually limit the number of iterations of RANSAC at training time (the pool of hypothesis) to just 20, which is much less than what is used at test time, creating a discrepancy between theory and practice.
We propose instead that the importance scores should be learned to minimize the expected loss of a randomly drawn sample:
\begin{align}\label{OP1}
    \textstyle \E_{\rm data} \Big[  \E_h \big[{\rm loss}({\rm solver}(h)) \big] \Big],
\end{align}
where the first expectation is over the training examples from the dataset and the second one is over a randomly chosen hypothesis. 
This loss is better aligned with the goal of RANSAC,  maximizing the probability of drawing good hypotheses and, thus, triggering the termination criterion in test time earlier. 
Additionally, this helps achieve a stable learning signal. 
In the remainder of this section, we will focus on estimating the gradient of~\eqref{OP1} in the scores $s$.

\subsection{Gumbel Softmax Sampler}
\label{sec:sampler}
The inner expectation in~\eqref{OP1} can be exactly computed in $\mathcal{O}(\binom{N}{k})$ time which is prohibitive in practice. 
Thus, at training time, we sample a mini-batch of hypotheses per image pair and perform one SGD step for this min-batch. However, by taking a discrete sample, the dependence on the parameters of the sampling distribution is lost. The expectation over hypotheses requires more careful consideration.

Provided that the solver and loss are differentiable, we approximate the derivative of the expectation in scores $s$ in a fashion similar to Gumbel-Softmax relaxation for categorical variables~\cite{jang2016categorical,maddison2016concrete}. 
Sampling from the PL distribution can be equivalently achieved by sorting noisy scores as
\begin{align}
    \textstyle (i_1\dots i_k) = \topk(s + \gamma), \ \ \ \gamma_i \sim {\rm Gumbel}(0,1),
\end{align}
where $\topk$ returns the indices of the largest $k$ elements and  ${\rm Gumbel}(0,1)$ is the standard Gumbel distribution.
We follow the straight-through GS strategy~\cite{jang2016categorical} to draw discrete samples on the forward pass but to propagate back using the Jacobian of the softmax operator. Let $y_j$ be a one-hot encoded index $i_j$, the index of the $j$th element in the sorting order. 
We define its $N\times N$ Jacobian in scores $s$ as
\begin{align}\label{GS-ST}
    \textstyle  \frac{d y_j}{d s} := \frac{d \softmax (\tilde s /\tau)}{d s}, 
\end{align}
where $\tau$ is the ``temperature of the relaxation'' hyper-parameter.
Using the vector of one-hot top-k indices $Y = (y_1, \dots y_k)$, we can easily select respective geometric features by matrix-matrix product 
    $\textstyle h = Y \Phi$.
%
Assuming both the solver and the loss function to be differentiable,~\eqref{GS-ST} is sufficient to complete the chain rule. 
Note that backpropagation using \eqref{GS-ST} requires only $\mathcal{O}(N)$ time and not $\mathcal{O}(N^2)$. A convenient way to implement our GS sampler is as:
\begin{subequations}
\begin{align}
    & \tilde s = s + \gamma, \ \ \ \ \gamma_i \sim {\rm Gumbel}(0,1),\\
    & Y = \onehot(\topk(\tilde s)), \\
    & \tilde y = \softmax(\tilde s /\tau), \\
    & Y = Y + \tilde y - {\rm sg}(\tilde y), \\ \label{stopgrad}
    & h = Y \Phi,
\end{align}
\end{subequations}
where ${\rm sg}$ does not propagate gradient (\ie, $\rm detach$ in PyTorch). 
Step \eqref{stopgrad} is a common trick: on the forward pass, the value equals precisely to $Y$, the one-hot indices of a correct sample. 
On the backward pass, the gradient flows through $\tilde y$ only. This sampler is visualized in Fig.~\ref{fig:gumbel_sampler}.

Let us remark that other differentiable samplers for PL distribution can also be applied, particularly NeuralSort~\cite{NeuralSort}.  
The proposed sampler is faster and performs well in our experiments even with the default $\tau = 1$. 
Furthermore, other methods can also be applied, such as unbiased {\sc reinforce}~\cite{Williams1992} with a relaxation-based baseline~\cite{Gadetsky-20}. We leave such refinements to future work. 
Note that all these options require the solver to be differentiable.

\subsection{Differentiable Solver}\label{sec:diff_solver}

Geometric solvers are a fundamental part of RANSAC-like approaches. Most solvers commonly used in computer vision can be made differentiable by implementing them in an automatic differentiation framework such as Pytorch and carefully considering each algorithmic component.

Learning-based pruning~\cite{clnet2021} propagates gradients through the well-known normalized eight-point (8PC) algorithm~\cite{hartley1997defense} for estimating essential (\textbf{E}) or fundamental (\textbf{F}) matrices.
There are two practical issues with the 8PC algorithm.
First, and most importantly, the 8PC and 7PC solvers have a degeneracy when the points stem from a close-to-planar 3D structure.
In this case, a degenerate model is estimated that, while often having a large number of inliers~\cite{chum2005two}, is incorrect w.r.t.\ the scene geometry.  
This misguides the learning in scenes dominated by planar structures and deteriorates the performance. 
Second, using eight correspondences instead of the minimal (5 for \textbf{E}, and 7 for \textbf{F} matrix) in RANSAC substantially decreases the chance to draw an all-inlier minimal sample and thus leads to a larger expected time to find a good solution or worse average quality of a solution found in a fixed budget. 

For \textbf{E} matrix estimation, numerous 5PC solvers have been developed~\cite{nister2004efficient,li2006five,batra2007alternative,stewenius2008minimal,bujnak2012making}.
However, practical applications (\eg, SfM~\cite{schonberger2016structure,theia-manual}) use either the method of Stewenius \etal~\cite{stewenius2008minimal} due to its stability, or that of Nister~\cite{nister2004efficient} due to its effectiveness. 
We use the method of Nister since it leads to more stable gradients in our experiments. 
Its steps are as follows: it creates the coefficient matrix from the input correspondences, 
decomposes it by SVD, solves a linear system of equations, solves a set of polynomial equations, and finally, basic arithmetic operations. 
We implemented a differentiable polynomial solver based on Sturm sequences~\cite{gonzalez1989sturm} and also one using companion matrices.
While both algorithms work well, the companion matrix-based solution we applied is the fastest. 

Fundamental matrix estimation is an easier problem, where the 7PC~\cite{hartley2003multiple} solver, besides an SVD decomposition on the coefficient matrix, only solves a cubic polynomial.
As we are given a closed-form solution for the cubic problem, we can straightforwardly make the entire algorithm differentiable. 
Note that, in our experiments, we have not observed improvements by replacing 8PC with 7PC.
This might be due to their shared degeneracy 
of planar scenes.

Minimal solvers often produce multiple solutions that all explain the data. 
While they are consistent with the constraints, most are inconsistent with the scene geometry.
At inference, the best one is selected based on its score. 
To mimic the test-time evaluation in training, the best algebraic solution is selected for each sample based on the model score and used for the loss computation. Respectively, the gradient propagates back through the best solution.

Note that we are not aware of public implementations of such solvers, neither in open-source libraries~\cite{van2014scikit,riba2020kornia} nor in standalone public repositories.
Therefore, we consider this a technical contribution to the community.

\subsection{Trainable Quality Function}
\label{sec:quality}

In RANSAC, the quality of an estimated model is calculated as the cardinality of its support, \ie, the inlier number.
Since RANSAC, a number of algorithms~\cite{torr2000mlesac,barath2019magsac,barath2020magsac++,barath2022learning} have improved the performance by better modeling the noise both in the inliers and outliers.
However, all such methods perform a best model selection step based on the maximal quality, which renders the procedure non-differentiable. 
Some works~\cite{Brachmann_2017_CVPR} tackle this problem by employing soft probabilistic hypothesis selection. 
Other methods~\cite{cne2018, clnet2021} combine classification loss with regression and geometry-induced losses~\cite{brachmann2019neural} to reason about the quality of a model. 

Instead of the above solutions, we exploit \textit{all} models $\{ \hat{\theta}_i \}_{i = 1}^t$ estimated during the fixed $t \in \mathbb{N}^{>0}$ iterations.
In each iteration, the estimated model is compared to the ground truth, and its implied loss is calculated, \eg, as the relative pose error in the case of $\mathbf{E}$ matrix estimation. Specifically, we consider the following loss measures.

In the case of relative pose estimation, the {\em pose error loss} is defined as follows. The solution $\hat \theta$ is decomposed into a rotation and translation $(\hat R, \hat t)$ using SVD~\cite{hartley2003multiple}. Then
\begin{equation}\label{eqn:loss_pose}
    \small
    \textstyle L_\text{pose} = \frac{1}{2} (\epsilon_\mathbf{R}(\hat{\mathbf{R}}, \mathbf{R}) + \epsilon_\mathbf{t}(\hat{\mathbf{t}}, \mathbf{t})),
\end{equation}
where $(\mathbf{R}, \mathbf{t})$ is the ground truth rotation and translation, and functions $\epsilon_\mathbf{R}$ and $\epsilon_\mathbf{t}$ compute the rotation and translation errors, respectively:
$\textstyle \epsilon_\mathbf{R}(\hat{\mathbf{R}}, \mathbf{R})= \cos^{-1} ( ( \text{tr} ( \hat{\mathbf{R}} \mathbf{R}\T) - 1 ) / 2)$, 
$\textstyle  \epsilon_\mathbf{t}(\hat{\mathbf{t}}, \mathbf{t}) = \cos^{-1} (\hat{\mathbf{t}}\T \mathbf{t}/ \| \hat{\mathbf{t}}\| \| \mathbf{t}\|)$.
The {\em average symmetric epipolar error} is defined as follows:
\begin{equation}
    \small
    L_\text{epi} = \frac{1}{|\mathcal{I}|} \sum_{i \in \mathcal{I}} \epsilon_\text{epi} (\theta, \Phi_i),
    \label{eqn:loss_epi}
\end{equation}
where $\mathcal{I}$ is the inlier set selected by the ground truth model $\theta$, $\Phi_i$ is the geometric features of a match $i$ and $\epsilon_\text{epi}$ is the respective residual error.
The overall loss minimized during training is the linear combination of the above ones as:
\begin{equation}
    \small
    L  = w_\alpha L_\text{pose} + w_\beta L_\text{epi},
    \label{eqn:overall}
\end{equation}
where $w_\alpha, w_\beta$ are weighting parameters.
The above-mentioned metrics are popular for evaluation and are differentiable. Since $\nabla$-RANSAC is differentiable, it enables a direct optimization on such evaluation metrics to learn the sampling probabilities and geometric features.

\subsection{Training and Testing Details}

The input of $\nabla$-RANSAC is a set of tentative correspondences obtained by any feature detector and matcher. 
The number of matches is fixed to 2K.
We choose the best 2K ones based on the matching score if we are given more.
In case of having fewer correspondences, we fill up the missing values with zeros.
We extract local and global features from the correspondences by a consensus learning block as backbone~\cite{clnet2021}.
We integrate over the extracted geometric information, \eg, scale, and orientation of the local descriptors, to help learn the qualities of correspondences.
\begin{figure*}[t]
    \centering
    \includegraphics[width=0.93\textwidth]{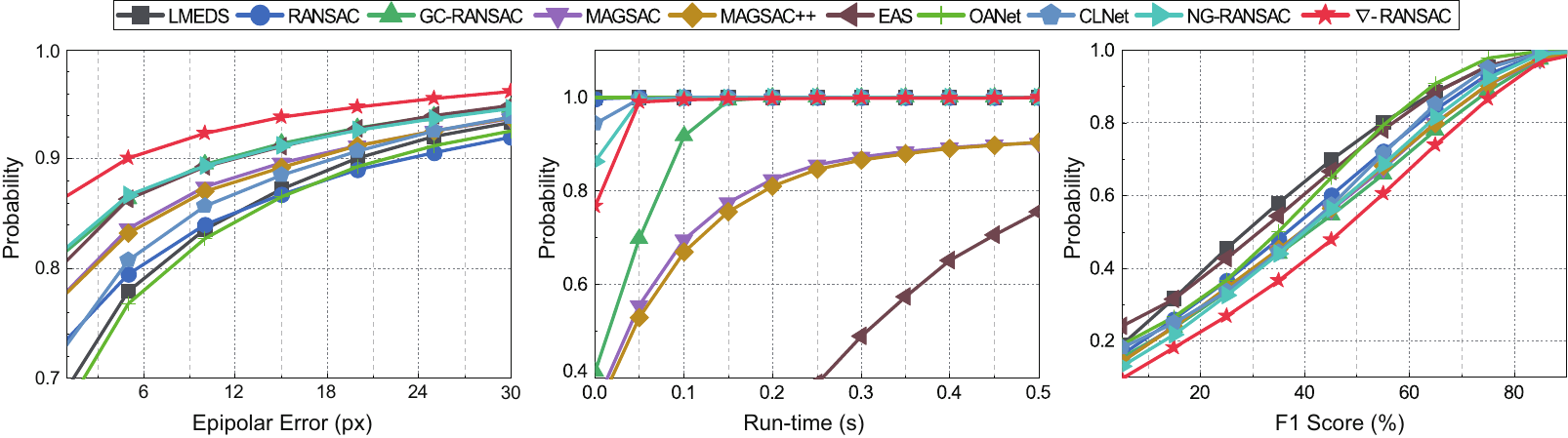}
    \vspace{0.5em}
    \caption{$\nabla$-RANSAC performance on 12 scenes of PhotoTourism measured by the cumulative distribution functions (CDF) of the epipolar errors (\textit{left; in pixels}), run-times (\textit{middle; in seconds}), and F1 score (\textit{right; in percentages}) for $\matr{F}$ matrix estimation.
    We use the thresholds as in \cite{barath2021marginalizing} for the traditional algorithms.
    Besides, OANet, CLNet, and NG-RANSAC were retrained on the same datasets as $\nabla$-RANSAC.
    In the left two plots, being close to the top-left corner indicates accurate results. 
    The bottom-right corner is preferable in the last plot. 
    }
    \label{fig:cdf}
\end{figure*}

\begin{table*}[t]
    \centering
    \small
    \setlength{\tabcolsep}{3pt}
  \resizebox{0.98\textwidth}{!}{\begin{tabular}{lcccccccccc}
      \toprule
Dataset / Method  &LMEDS~\cite{rousseeuw2005robust} & RSC~\cite{RANSAC} 
 &GC-RSC~\cite{barath2018graph}& MSC~\cite{barath2019magsac}& MSC++~\cite{barath2020magsac++} & EAS~\cite{fan2021efficient} & OANet~\cite{oanet2019} &CLNet~\cite{clnet2021} &NG-RSC~\cite{brachmann2019neural} & $\nabla$-RANSAC \\\midrule
  Avg. time (ms)~$\downarrow$ & $\textbf{21.00}$  & $30.67$  & $73.25$  & $281.3$ &$ 318.13$ & $325.83$ &$\textbf{21.00}$ &$34.83$& $\underline{25.85}$  & $28.66$  \\
  \midrule
  
  Buckingham Palace & $23.26$ & $24.95$ & $26.48$& $27.23$ & $26.28$  & $\underline{28.28} $& $24.70$ & $27.46$ &$28.06$&$\textbf{33.05} $ \\
  
  Brandenburg Gate & $31.74$ & $39.70$ & $43.01$& $42.02$ & $42.43$ &$34.84$ & $42.49$ & $39.69$ & $\underline{43.19}$& $\textbf{47.66}$\\ 
  
  Colosseum Exterior & $43.32$ & $48.25$ & $50.86$& $51.76$ & $51.56$ &$50.49$ & $40.50$& $47.12$ & $\underline{52.42}$ & $\textbf{55.80}$\\ 
  
  Grand Place Brussels & $27.45$ & $31.42$ & $\underline{33.31}$& $32.60$& $33.08 $& $32.40$&$29.10$ & $31.80$& $32.13$& $\textbf{35.61}$\\
  
  Notre Dame Front Facade & $30.39$ & $37.20$ & $40.48$& $39.35$ & $39.00$ &$39.98$& $33.17$ & $38.20$ & $\underline{40.59}$  & $\textbf{46.10}$\\
  
  Palace of Westminster & $22.20$ & $28.29$ & $33.15$& $31.94$ & $31.56$ & $32.42$ & $31.33$ & $32.96$ & $\underline{33.54}$ & $\textbf{41.15}$\\
  
  Pantheon Exterior & $54.22$ & $59.23$ & $\underline{62.26}$& $61.86$ & $61.60$ & $60.54$&$49.89$ &$56.81$& $61.31$ & $\textbf{64.48}$ \\
  
  Prague Old Town Square & $26.61$ & $30.14$ & $32.48$& $32.39$ & $31.30$ & $34.35$& $34.05$ & $\underline{37.58}$ & $33.13$& $\textbf{37.80}$\\
  
  Sacre Coeur & $41.58$ & $49.03$& $56.36$ & $53.23$ &$53.06$ & $45.10$ & $41.30$ &$45.09$ & $\underline{56.61}$ &$\textbf{61.52}$\\ 
  
  Taj Mahal & $38.43$ & $48.44$ & $51.51$& $50.71$ & $50.43$ &$51.63$ & $50.04$ &$52.17$ &$\underline{54.71}$ & $\textbf{58.58}$\\
  
  Trevi Fountain & $29.85$ & $31.67$ & $\underline{34.99}$& $33.94$ & $34.28$ &$33.75$& $27.69$ &$30.82$ & $34.61$& $\textbf{39.11}$\\
  
  Westminster Abbey & $50.97$ & $52.27$ & $\underline{55.99}$& $55.15$ & $54.91$ & $53.07$ &$42.10$ & $48.37$ & $53.61$ &$\textbf{56.55}$\\ 
  \cline{0-0}
  
  Avg.\ over all scenes~$\uparrow$ & $35.00$ & $40.10$ & $43.41$ & $42.68$ & $42.46$ & $42.91$& $36.91$ & $40.67$ & $\underline{43.66}$& $\textbf{48.12}$ \\ \bottomrule\\
    \end{tabular} }
    \caption{
The average run-time (\textit{first row; in milliseconds})  and F1 scores (\textit{each, average in last row}) for \textbf{F} matrix estimation on 12K image pairs from the PhotoTourism dataset~\cite{IMC2020}. 
  We use the threshold tuned in~\cite{barath2021marginalizing} for RANSAC, GC-RANSAC, MAGSAC, and MAGSAC++. 
  We tuned the parameters of EAS, and retrained OANet, CLNet, NG-RANSAC on the same data as training $\nabla$-RANSAC.
  The results with the pre-trained models provided by the authors are in Tab.~\ref{tab:pretrained_clnet_compare_f}. 
  Best results are \textbf{bold}, the second best \underline{underlined}.
    }
  \label{tab:photo_result_f}
  \vspace{-1em}
\end{table*}

\paragraph{Initialization.}
We apply a 1K epoch-long weight initialization procedure as in~\cite{brachmann2019neural}. For this initialization, the KL divergence between the predicted importance ${\rm Cat}(p)$ and the target categorical distributions is minimized. The target distribution is chosen proportional to $\softmax$ of residuals of all points from the GT model~\cite{brachmann2019neural}. This initialization scheme does not require sampling the model hypothesis.

\paragraph{Training.}
Along with the initialized weights, the gradient clipping~\cite{chen2020understanding} technique is used to avoid exploding gradients, make the training stable and accelerate the convergence. 
In the \textbf{F} matrix case, we normalize the points for consensus learning and use the original unnormalized ones in minimal solvers. 
We train the pipeline using the 8PC and 7PC algorithms for \textbf{F} estimation with fixed 1K iterations.
For \textbf{E} case, we use the 5PC algorithm and 100 iterations.

\paragraph{Testing.}
At test time, we equip state-of-the-art RANSAC components.
The model trained end-to-end provides importance scores to the weighted random sampler. Drawing a sample from PL distribution with scores $s$ is simplified as:
\begin{align}
    \textstyle (i_1\dots i_k) = \topk(u_i^{1/s_i}  ), \ \ \ u_i \sim {\rm Uniform}(0,1).
\end{align}
We use the MAGSAC++~\cite{barath2020magsac++} model quality function to select the best model, marginalizing over a range of noise scales. 
We also apply an inner RANSAC-based local optimization~\cite{lebeda2012fixing} to improve the accuracy further.
Also, we perform the Levenberg-Marquardt~\cite{more1978levenberg} numerical optimization minimizing the pose error on all inliers as a final step.
The test code runs in C++ to be fast. 
\section{Experimental Results}
\label{sec:results}

Our main experiments for epipolar geometry estimation were conducted on 13 scenes from the training set of the CVPR IMW 2020 PhotoTourism benchmark \cite{IMC2020} that provides images, intrinsic and extrinsic camera parameters from a reference COLMAP reconstruction, and pre-detected RootSIFT features~\cite{RootSIFT2012}.  
We train and validate the method on scene St.\ Peter's Square consisting of 4950 image pairs, split $3$ to $1$ into training and validation sets.
The other 12 scenes are used for testing.
We compare $\nabla$-RANSAC on fundamental (\textbf{F}) and essential (\textbf{E}) matrix estimation to classical robust estimators, \ie, RANSAC~\cite{RANSAC}, LMEDS~\cite{rousseeuw2005robust}, and their recent alternatives, such as GC-RANSAC~\cite{barath2018graph}, MAGSAC~\cite{barath2019magsac}, MAGSAC++~\cite{barath2020magsac++} and EAS~\cite{fan2021efficient}. 
Also, we test the provided models by the state-of-the-art learning-based methods, OANet~\cite{oanet2019}, CLNet~\cite{clnet2021} (both followed by RANSAC), NeFSAC~\cite{cavalli2022nefsac}, and NG-RANSAC~\cite{brachmann2019neural}.
To make fair comparison, we retrained NG-RANSAC~\cite{brachmann2019neural}, CLNet~\cite{clnet2021}, and OANet~\cite{oanet2019} on the same data as what we train $\nabla$-RANSAC on. 
Moreover, we train and evaluate on ScanNet~\cite{dai2017scannet} following the widely used 
 feature matcher SuperGlue~\cite{sarlin2020superglue}. 
In addition, we apply the proposed method to point cloud registration by training and testing GeoTransformer~\cite{qin2022geometric} features of 3DMatch~\cite{zeng20173dmatch} and 3DLoMatch~\cite{huang2021predator}, shown in Sec.~\ref{sec:rigid}.
\paragraph{Technical details.}
There are two setups of training with $\nabla$-RANSAC. 
\textit{First}, we train the consensus learning module~\cite{clnet2021} with off-the-shelf features: \textbf{Outdoors} is trained with the SNN ratio~\cite{lowe1999object} coming from RootSIFT descriptors, and the underlying feature scales and orientations as learnable side information besides coordinates, where prefiltering (threshold=0.8) and initialization are needed; \textbf{Indoors} is trained with the coordinates and confidence output from the most commonly used feature detector and matcher, \ie, SuperPoint~\cite{detone2018superpoint} with SuperGlue~\cite{sarlin2020superglue}.
For these two cases, we use $0.75$ and $3.0$ pixels as the inlier-outlier thresholds for robust estimators, respectively.
\textit{Second}, a clearer setup of connecting $\nabla$-RANSAC directly to learnable feature matching method, \ie, LoFTR~\cite{Sun_2021_CVPR}, to improve matching prediction with reliable confidence scores (Section~\ref{subsec:matching}).
All the experiments were conducted on Ubuntu 20.04 with GTX 3090Ti, OpenCV 4.5.5, and PyTorch 1.11.1 with Cuda 11.3.1. 
We re-implemented RANSAC components in PyTorch and connected them with other modules for training. 

\subsection{Fundamental Matrix Estimation}
Benefiting from the proposed $\nabla$-RANSAC, we trained the model parameters jointly with the prediction network to learn the statistical features of tentative matches and predict inlier probabilities. 
We adopt one consensus learning block from~\cite{clnet2021} but without filtering the points with their predicted probabilities. 
The model is trained for 10 epochs and optimized by Adam~\cite{kingma2014adam} with a learning rate of $1e^{-5}$.
For \textbf{F} estimation, 
the coordinates are normalized by the image sizes before training.
For testing, we use 1K randomly chosen image pairs from each of the remaining \num{12} scenes. 
Thus, the methods are tested on 12K image pairs in total.
To measure the quality of the estimated \textbf{F} matrices, we use the F1 score in percentage (\%) and median epipolar errors in pixel (px)following~\cite{brachmann2019neural}.
We use the normalized 8PC algorithm for training as it leads to more stable solutions than the 7PC solver in our experiments.

The average F1 scores and the run-time of the robust estimation (in milliseconds) are reported in Tab.~\ref{tab:photo_result_f} on each scene of the dataset and also averaged overall in the last row. 
The proposed $\nabla$-RANSAC achieves the most accurate results on all but one scene, where it is the second-best by a small margin. 
On average, it improves by ${\sim}5$\% compared to the second best method. 
$\nabla$-RANSAC runs at a comparable speed to less accurate alternatives.
The most efficient method is LMeDS achieving an 11.61\% lower F1 score than $\nabla$-RANSAC.
Note that these timings do not include the inference time, around $2$ ms on average using GPU. 
Also, we show the comparison of the proposed method with the given pre-trained models of the state-of-the-art learning-based robust estimators in Tab.~\ref{tab:pretrained_clnet_compare_f}. 
We perform better in terms of F1 scores and errors, with comparable run-time, even though we trained on the least data among the methods in the table.
Also, we test on the provided models by the recent work, NeFSAC~\cite{cavalli2022nefsac}.
The proposed method leads to significant improvements compared to NeFSAC. 
It is important to note, however, that the contributions of $\nabla$-RANSAC are complementary to that of NeFSAC. Thus, they can be straightforwardly combined together. 

Furthermore, we show the cumulative distribution functions (CDF) of the epipolar errors, run-times, and F1 scores calculated from the 12K test image pairs in Fig.~\ref{fig:cdf}.
In the two left plots of epipolar error and run times, being close to the top-left corner indicates better performance.
The epipolar error curve of the proposed $\nabla$-RANSAC is above the other methods on the entire plot.
While not the fastest, it finishes in $0.1$ seconds in $100$\% of the test cases.
This confirms that $\nabla$-RANSAC is applicable in time-sensitive applications. 
The right figure shows the CDFs of the F1 scores. 
In contrast to the other plots, being close to the bottom-right corner is preferred. 
$\nabla$-RANSAC has a better score on the entire range.
It coincides with other methods only at the end of the range, as supposed to end up with $1$. 

\begin{table}[t]
    \centering
    \setlength{\tabcolsep}{3pt}
    \resizebox{0.99\columnwidth}{!}{\begin{tabular}{lccccc}
    \toprule 
    Metric / Method & OANet~\cite{oanet2019} & CLNet~\cite{clnet2021} & NeFSAC~\cite{cavalli2022nefsac}&NG-RSC~\cite{brachmann2019neural} &$\nabla$-RANSAC  \\
    \midrule
    F1 Score (\%)~$\uparrow$ & $42.29$ & $38.61$ & $43.17$&$45.80$ & $\textbf{48.12}$\\
   Med.\ epi.\ error (px)~$\downarrow$ & \phantom{1}$2.50$ & \phantom{1}$8.73$ &$1.92$ & \phantom{1}$1.51$ &\phantom{1}$\textbf{0.86}$ \\
    Time (ms)~$\downarrow$ & $\textbf{21.75}$ &$ 27.83$ & $33.58$ & $25.90$ &$28.66$ \\
    \bottomrule\\
    \end{tabular}}
    \caption{Comparison of $\nabla$-RANSAC and the models provided by the authors of NG-RANSAC, OANet, CLNet and the recent work, NeFSAC for \textbf{F} matrix estimation on PhotoTourism. CLNet and OANet were trained on more than 541K image pairs from YFCC~\cite{thomee2016yfcc100m}. Note that we train on \num{4950} image pairs of one specific scene and show good generalization on real-world data. NG-RANSAC uses twice as many imag pairs as us.
    }
    \label{tab:pretrained_clnet_compare_f}
    \vspace{-1em}
\end{table}

\vspace{1mm}
\noindent
\subsection{Essential Matrix Estimation}
We evaluate \textbf{E} matrix estimation both on RootSIFT features of PhotoTourism~\cite{snavely2006photo} (outdoors) as used for \textbf{F} estimation, and SuperPoint features matched with SuperGlue on ScanNet~\cite{dai2017scannet} (indoors).
The correspondences are normalized by the intrinsic matrices. 
\textbf{E} matrix estimation is trained with our differentiable 5PC solver.
for 10 epochs, where the iteration number of robust estimation is fixed to 100.
For evaluation, we decompose the \textbf{E} matrix to rotation and translation, calculate their errors $\epsilon_\mathbf{R}$, $\epsilon_\mathbf{t}$ and report the maximum of rotation and translation errors $\max(\epsilon_\mathbf{R}, \epsilon_\mathbf{t})$.
We calculate the Area Under the Recall curve (AUC) thresholded at $5^\circ$, $10^\circ$, and $20^\circ$ following the previous work~\cite{cne2018, brachmann2019neural}. 

\paragraph{PhotoTourism~\cite{IMC2020}.}
The AUC scores averaged over 12 testing scenes from PhotoTourism are reported in Tab.~\ref{tab:e}.
The highest AUC scores, at all thresholds, are achieved by $\nabla$-RANSAC. For example, its AUC@$5^\circ$ score is higher than that of the second best methods (\ie, MAGSAC and MAGSAC++) by AUC $3$ points. 
It has comparable run-time to other less accurate alternatives. 

Instead of comparing with retrained models, we ran the pretrained CLNet~\cite{clnet2021} and OANet~\cite{oanet2019} models provided by the authors for \textbf{E} estimation on the test scenes as used in \cite{barath2022learning}. Both methods finish with a RANSAC on their inliers. 
The results in Tab.~\ref{tab:clnet} demonstrate that $\nabla$-RANSAC achieves considerably better results even when trained on a fraction of data used for other methods. 
\begin{table}[t]
    \centering
    \setlength{\tabcolsep}{3pt}
    \resizebox{0.83\columnwidth}{!}{\begin{tabular}{lccccc}
    \toprule 
    Method & AUC@$5^\circ~\uparrow$ & AUC@$10^\circ~\uparrow$ & AUC@$20^\circ~\uparrow$& Time (ms)~$\downarrow$\\ \midrule
    LMEDS \cite{rousseeuw2005robust} & $0.24$ & $0.30$ & $0.37$ & $\textbf{\phantom{1}27}$\\
    RANSAC~\cite{RANSAC}& $0.26$ & $0.32$ & $0.40$ & $\phantom{1}88$\\
    GC-RANSAC~\cite{barath2018graph} & $0.33$ & $0.37$ & $0.42$ & $175$\\
    MAGSAC~\cite{barath2019magsac} & $0.37$ & $0.42$ & $0.47$ & $239$\\
    MAGSAC++~\cite{barath2020magsac++}& $0.37$ & $0.42$ & $0.47$ & $113$\\
    EAS~\cite{fan2021efficient} & $0.24$ & $0.28$ & $0.34$ & $325$\\
    OANet~\cite{oanet2019}& $0.29$ & $0.33$ & $0.39$ & $\phantom{1}49$\\
    CLNet~\cite{clnet2021}& $0.34$ & $0.40$ & $0.47$ & $\phantom{1}58$\\
    NeFSAC~\cite{cavalli2022nefsac} & 0.34 & 0.40 & 0.45 &374\\
    NG-RSC~\cite{brachmann2019neural} & $0.35$ & $0.41$ & $0.47$ & $\phantom{1}80$\\
    $\nabla$-RANSAC & $\textbf{0.41}$ & $\textbf{0.45}$ & $\textbf{0.50}$ & $117$\\
    \bottomrule\\
    \end{tabular}}
    \caption{The average AUC scores of $\nabla$-RANSAC and comparison methods over 12 scenes from PhotoTourism, under different thresholds.  We are the most accurate method for \textbf{E} estimation.
    \vspace{-0.5em}
    }
    \label{tab:e}
\end{table}

\begin{table}[t]
    \centering
 \resizebox{0.99\columnwidth}{!}{
    \begin{tabular}{c|c c c c}
    \toprule
       Method  & train data & AUC@$10^\circ~\uparrow$  & med. \textbf{R} error~$\downarrow$   & med. \textbf{t} error~$\downarrow$ \\
       \midrule
       pretr.\ OANet~\cite{oanet2019} & YFCC~\cite{thomee2016yfcc100m}, 541K  & 0.67 & 2.12 &5.26 \\
       pretr.\ CLNet~\cite{clnet2021} & YFCC~\cite{thomee2016yfcc100m}, 541K  & 0.69 & 1.75 &4.34 \\
       $\nabla$-RANSAC &  St.\ Peter's, 55K & \textbf{0.77} & \textbf{1.26} & \textbf{2.91}\\
        \bottomrule
    \end{tabular}}\vspace{2mm}
\caption{\textbf{E} estimation performance of the proposed method and the pretrained CLNet and OANet, both finishing with a RANSACas a post processing procedure, on the testing scenes used in \cite{barath2022learning}.}
    \label{tab:clnet}
    \vspace{-1em}
\end{table}

\paragraph{Other Scenes from PhotoTourism as in \cite{barath2022learning}.}
In \cite{barath2022learning}, the authors compare on the test set that consists of entirely different scenes from the last paragraph.
To be comparable to MQNet~\cite{barath2022learning}, we report results on this split in Tab.~\ref{test_on_test_scenes}. 
The proposed $\nabla$-RANSAC leads to substantial improvements compared to MQNet~\cite{barath2022learning} and MAGSAC++~\cite{barath2020magsac++}, both in terms of rotation (\textbf{R}) and translation (\textbf{t}) matrix estimation accuracy.  
\begin{table}[h]
    \centering
 \resizebox{0.90\columnwidth}{!}{
    \begin{tabular}{c|c|c c c}
    \toprule
      Task    &AUC@$10^\circ~\uparrow$ & MAGSAC++~\cite{barath2020magsac++} & MQNet~\cite{barath2022learning} & $\nabla$-RANSAC \\
      \midrule
\textbf{E} est.  & \textbf{R} / \textbf{t} & 0.71 / 0.47  &0.79 / 0.62  &\textbf{0.84 / 0.74}\\
       \textbf{F} est. & \textbf{R} / \textbf{t} & 0.64 / 0.31 & 0.70 / 0.36  &\textbf{0.79 / 0.53}\\
      \bottomrule
        \end{tabular}}\vspace{2mm}
\caption{Rotation and translation estimation performance on the testing scenes from PhotoTourism~\cite{snavely2006photo} as used in MQNet~\cite{barath2022learning}.}
    \label{test_on_test_scenes}
    \vspace{-1em}
\end{table}
\begin{table}[h]
    \centering
    \resizebox{0.99\columnwidth}{!}{\begin{tabular}{l ccccc}
    \toprule
         Method & Confidence&AUC@$5^\circ~\uparrow$ &AUC@$10^\circ~\uparrow$ &AUC@$20^\circ~\uparrow$ &run-time ($\mu$s)~$\downarrow$\\
         \midrule 
         RANSAC &- & $16.02$ & $33.53$ & $51.84$  & $110.2$\\ 
         MAGSAC++ &- & $17.70$ & $35.15$ & $51.75$ & $\phantom{1}58.9$\\
         MAGSAC++&SuperGlue & $18.67$ & $35.85$ & $52.60$ & $\phantom{1}51.5$\\
         MAGSAC++ &$\nabla$-trained & $\textbf{19.15}$ & $\textbf{36.40}$ & $\textbf{53.47}$ & $\phantom{1}\textbf{49.3}$ \\

        \bottomrule \\
    \end{tabular}}
    \caption{\textbf{E} matrix evaluation on 1500 test pairs of ScanNet used in SuperGlue~\cite{sarlin2020superglue}. The last two rows are tested on MAGSAC++ with PROSAC sampler guided by the confidence predicted from the provided SuperGlue model and $\nabla$-RANSAC trained on SuperGlue matches. Note the run-time is in \textit{microseconds}.}
    \label{tab:scannet}

\end{table}

\paragraph{ScanNet~\cite{dai2017scannet}.}
Following the matching and evaluation process in SuperGlue~\cite{sarlin2020superglue}, we trained and tested $\nabla$-RANSAC with the most popular indoor 3D point cloud dataset, \ie ScanNet, consisting of 1201 scans for training and 312 scans for validation. 
The tentative correspondences are detected and matched using SuperPoint~\cite{detone2018superpoint} and SuperGlue.
We train $\nabla$-RANSAC using 16790 randomly selected pairs from 1201 scans and 4680 for validation.
The 1500 testing pairs are the same ones used in the SuperGlue paper. 
We evaluate $\nabla$-RANSAC by comparing the AUC scores of RANSAC and MAGSAC++, PROSAC sampling with either SuperGlue confidence or $\nabla$-RANSAC prediction.
As shown in Tab.~\ref{tab:scannet}, our trained weights work better in guided sampling than the confidence given by SuperGlue. 

\subsection{3D Point Rigid Registration}
\label{sec:rigid}

To demonstrate that the proposed $\nabla$-RANSAC is general, we apply to rigid point cloud registration.
As the differentiable minimal solver, we implemented the standard Kabsch algorithm~\cite{kabsch1976solution} in PyTorch.  
We trained $\nabla$-RANSAC on GeoTransformer~\cite{qin2022geometric} correspondences detected in the 3DMatch~\cite{zeng20173dmatch} and 3DLoMatch~\cite{huang2021predator} benchmarks.
We use 75 scenes from 3DMatch dataset, 14k scenes in total for training, 8 scenes for validation (1331 pairs), and 8 for testing (1623 pairs) following \cite{qin2022geometric}. 
In addition, another 8 scenes from 3DLoMatch are used for testing with 1781 scenes included.
Note that 3DLoMatch is particularly challenging due to the low overlap, \ie, below 30\%.

To measure the error of the registration, we calculate the Relative Rotation Error (RRE), Relative Translation Error (RTE), and Root Mean Squared Error (RMSE), Registration Recall (RR). 
We optimize the CLNet~\cite{clnet2021} inlier probability predictor by the proposed $\nabla$-RANSAC.
We then feed MAGSAC++~\cite{barath2020magsac++} with PROSAC the original GeoTransformer match confidence predictions and the ones optimized by the proposed method. 

As shown in Tab.~\ref{tab:3dlomatch}, \ref{tab:3dmatch}, the inlier priors predicted by $\nabla$-RANSAC better guide the PROSAC sampler than using the original confidences directly from GeoTransformer. 
\begin{table}[h]
\vspace{-0.5mm}
    \hspace{2mm}\resizebox{0.95\columnwidth}{!}{\begin{tabular}{ c  c|  c c  c c }
    \hline   
       Inlier prob.\ predictor &\# iters & RRE ($^\circ$)~$\downarrow$ &RTE (cm)~$\downarrow$ & RMSE (cm)~$\downarrow$ & RR ($\%$)~$\uparrow$\\
    \hline   
        \multirow{3}{*}{GeoTransformer~\cite{qin2022geometric}} & \phantom{0}1K &27.91&66.64&28.36&76.55\\
        & 10K &27.78&68.48&28.41& 75.78\\
         & 50K&27.76&67.38&28.91 &76.36\\ \hline
       \multirow{3}{*}{$\nabla$-RANSAC}  & \phantom{0}1K& 26.25&64.57&\textbf{27.21}&\textbf{77.23}\\
        & 10K &\textbf{25.72}&\textbf{62.34}&27.55&76.94\\
        & 50K & 25.99&63.81 &27.45&76.94 \\
    \hline   
\end{tabular}}\vspace{2mm}
\caption{Point cloud registration using MAGSAC++ with PROSAC sampler on GeoTransformer matches on the 3DLoMatch~\cite{huang2021predator} dataset. 
The inlier probabilities used inside PROSAC are the ones that GeoTransformer outputs (top rows), or learned by the proposed $\nabla$-RANSAC (bottom). The best results are in \textbf{bold}.}
\label{tab:3dlomatch}
\vspace{-1mm}
\end{table}

\begin{table}[h]
    \hspace{2mm}\resizebox{0.95\columnwidth}{!}{\begin{tabular}{ c  c|  c c  c c }
    \hline   
       Inlier prob.\ predictor &\# iters & RRE ($^\circ$)~$\downarrow$  &RTE (cm)~$\downarrow$& RMSE (cm)~$\downarrow$ & RR ($\%$)~$\uparrow$\\
    \hline   
        \multirow{3}{*}{GeoTransformer~\cite{qin2022geometric}} & \phantom{0}1K &5.76&15.80&6.67&96.90\\
        & 10K &6.31&16.99&6.97&96.43\\
         & 50K&6.27&16.42&6.80&96.55\\ \hline
       \multirow{3}{*}{$\nabla$-RANSAC}  & \phantom{0}1K& 5.55&14.68&6.53&\textbf{97.01}\\
        & 10K &\textbf{5.44}&\textbf{14.24}&\textbf{6.47}&96.90\\
        & 50K & \textbf{5.44}&14.45&6.57&\textbf{97.01}\\
    \hline   
\end{tabular}}\vspace{2mm}
\caption{Point cloud registration using MAGSAC++ with PROSAC on GeoTransformer matches on the 3DMatch~\cite{zeng20173dmatch} dataset. 
The inlier probabilities used inside PROSAC are the ones that GeoTransformer outputs (top rows), or learned by the proposed $\nabla$-RANSAC (bottom). The best results are in \textbf{bold}.}
\label{tab:3dmatch}
\vspace{-1em}
\end{table}
\subsection{Ablation Studies}
\label{subsec:ablation}

\paragraph{Objective functions.}
Tab.~\ref{tab:dsac_compare} compares training objectives optimized by $\nabla$-RANSAC running the same components (including Gumbel Sampler) in \textit{all} cases. 
The first row shows the results with the proposed Eq.~\ref{OP1} as the objective. 
The next two rows replace Eq.~\ref{OP1} with other objectives, \eg, the probabilistic selection approach of DSAC~\cite{Brachmann_2017_CVPR}. 
In the last row, we use REINFORCE for gradient calculation and backpropagation as \cite{brachmann2019neural} does.

For \textbf{F} estimation, $\nabla$-RANSAC trained with  Eq.~\ref{OP1} performs the best in terms of median epipolar error. SoftAM achieves a comparable F1 score but low efficiency. 
In addition, $\nabla$-RANSAC is the best in terms of \textbf{E} estimation accuracy with the second-best run-time, while the fastest REINFORCE~\cite{brachmann2019neural} achieves lower accuracy. 
Note that we cannot directly compare with DSAC, but the learning objective of DSAC can be integrated in our training as an option.

\begin{figure}[t]
    \centering
    \includegraphics[width = 0.95\columnwidth]{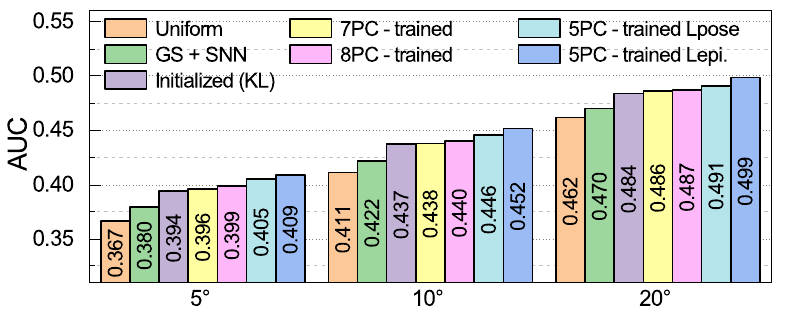}
    \caption{
    Ablation Studies of Gumbel Softmax (GS) sampler.
    AUC scores for \textbf{E} estimation using Uniform sampler, GS with SNN ratio, and learned weights (Initialized, 7PC-trained, 8PC-trained, and 5PC-trained) on 12K images from PhotoTourism.
}
    \label{fig:technical}
    \vspace{-0.5em}
\end{figure}

\paragraph{Differentiable Sampler.}
The sampling procedures are tested using MAGSAC++ on 12K image pairs from PhotoTourism in Fig.~\ref{fig:technical}.
Uniform and GS + SNN~\cite{lowe1999object} are uniform random sampling and GS guided by SNN ratios, respectively.
The other tested methods are GS with different weights: initialized (with KL divergence), trained with 7PC, 8PC, and 5PC with different losses. 
Epipolar error works better than pose error for \textbf{E} estimation. 
Compared with a uniform sampler and non-learnable weights for the GS sampler, $\nabla$-RANSAC weights perform better.

\paragraph{Differentiable Solvers.}
For \textbf{F} matrix estimation, 
the average results when $\nabla$-RANSAC is trained with the 7PC and norm.\ 8PC solvers are shown in Tab.~\ref{tab:F_solvers}.
For \textbf{F} estimation, the 8PC solver leads to the most accurate results while being the fastest.
As shown in Fig.~\ref{fig:technical}, the highest AUC scores at all thresholds for \textbf{E} matrix estimation are achieved by the 5PC method, confirming the necessity of using better minimal solvers for training rather than the 8PC algorithm.

\begin{table}[t]
    \centering
    \resizebox{0.99\columnwidth}{!}{\begin{tabular}{l|c c c| c c}
    \toprule
     \multirow{2}{*}{Learning Objectives}&\multicolumn{3}{c}{\textbf{F} matrix estimation} &\multicolumn{2}{|c}{\textbf{E} matrix estimation} \\
&F1 score~$\uparrow$ & med. epi. error~$\downarrow$  &run-time (ms)~$\downarrow$& AUC@10$^\circ~\uparrow$ &run-time (ms)~$\downarrow$ \\
    \midrule
Eq.~\ref{OP1} & \underline{48.12} & \textbf{0.86} & \textbf{28.66}  & \textbf{0.45} & \textbf{116.90}\\
        Prob. selection~\cite{Brachmann_2017_CVPR}  & 40.95 & 4.92 & \underline{34.21} & 0.43 & 134.48 \\
         SoftAM~\cite{Brachmann_2017_CVPR} & \textbf{48.17} & \underline{0.88} & 47.46 & \underline{0.44} & 147.88 \\
        REINFORCE~\cite{brachmann2019neural} & 42.77 & 4.06 & 34.75 & \underline{0.44} & \underline{125.65} \\
        \bottomrule 
    \end{tabular}}
 \vspace{0.5em}
    \caption{
    Performance of $\nabla$-RANSAC trained with different learning objectives:
    SoftAM and probabilistic model selection (DSAC) from \cite{Brachmann_2017_CVPR}; and the REINFORCE gradient approximation~\cite{brachmann2019neural} for \textbf{F} and \textbf{E} estimation.
    Best results in \textbf{bold}, the second best \underline{underlined}.}
    \label{tab:dsac_compare}
    \vspace{-1em}
\end{table}

\begin{table}[t]
    \centering
    \resizebox{0.83\columnwidth}{!}{\begin{tabular}{lccc}
    \toprule
    \textbf{F} matrix estimation &Initialized & 7PC~\cite{hartley2003multiple}-trained & 8PC~\cite{hartley2003multiple}-trained  \\  \midrule 
         F1 score (\%)~$\uparrow$ & $45.71$ & $46.04$ & $\textbf{47.21}$ \\ 
         Med.\ epi.\ error (px)~$\downarrow$ & $\phantom{1}1.73$ &  $\phantom{1}1.54$ & $\phantom{1}\textbf{1.00}$  \\
         \bottomrule \\
    \end{tabular}}
    \caption{Performance of $\nabla$-RANSAC trained with  \textbf{differentiable solvers} on \textbf{F} estimation, evaluated by the proposed GS sampler.
    The same 12 testing scenes are used from PhotoTourism~\cite{IMC2020}.}
    \label{tab:F_solvers}
    \vspace{-1em}
\end{table}

\subsection{Learning Feature Matching with $\nabla$-RANSAC}
\label{subsec:matching}
In this section, we tune an end-to-end feature matcher, LoFTR~\cite{Sun_2021_CVPR}, on the epipolar error using $\nabla$-RANSAC.
An advantage of our method is that it allows the RANSAC pipeline to be optimized for test-time metrics such as pose and epipolar errors. 
Even though they are differentiable themselves, their input comes from RANSAC. 
If RANSAC is non-differentiable, these metrics cannot be directly used as a loss function. 
Additionally, the entire feature computation and matching module can be directly optimized on such metrics if the RANSAC is differentiable. 

Following~\cite{patel2020learning,patel2021feds}, LoFTR was initialized with the weights from the standard LoFTR and trained.
Training of LoFTR with $\nabla$-RANSAC for \textbf{F} matrix estimation was performed on scene St.\ Peter's Square  of the PhotoTourism dataset, while the remaining $12$ scenes were used for testing as the main experiments. 
The training used AdamW optimizer~\cite{loshchilov2017decoupled} for $10$ epochs, with a learning rate of $1e^{-6}$. 
Note that CLNet was not used for this setup as LoFTR directly predicts the filtered correspondences and their confidence scores. 
Using the $\nabla$-RANSAC, the gradients of the loss are obtained with respect to both the correspondences and confidence scores. Unlike the main experiments, here we use the top-30\% of the estimated models for the training.

The evaluation is performed using three different inference protocols, namely OpenCV-RANSAC~\cite{RANSAC}, OpenCV-PROSAC~\cite{chum2005matching}, and MAGSAC-PROSAC~\cite{barath2020magsac++}, the results are presented in Tab. \ref{tab:loftr}. These evaluations use a threshold of $3$ pixels to determine a prediction as an inlier based on the epipolar error, different from the other experiments in the paper that use a threshold of $0.75$ px. The results show that the proposed $\nabla$-RANSAC can be used to improve learning-based feature-matching approaches. This observation indicates the robustness of $\nabla$-RANSAC as a pre-trained model that can be used without any adaptation plug-and-play.

\begin{table}
    \centering
    \setlength{\tabcolsep}{3pt}
    \resizebox{0.99\columnwidth}{!}{\begin{tabular}{lccccc}
    \toprule 
     Inference Protocol & LoFTR ~\cite{Sun_2021_CVPR} &F1 score (\%)~$\uparrow$ & avg. epi. error (px)~$\downarrow$ & run-time (ms)~$\downarrow$ \\ \midrule
    \multirow{2}{*}{OpenCV-RANSAC~\cite{RANSAC}}  & Standard &  $64.07$ & $13.16$ & $\phantom{1}2.83$ \\
    & $\nabla$-trained & $\textbf{64.43}$ & $\textbf{11.90}$  & $\phantom{1}\textbf{2.81}$ \\ \hline
    \multirow{2}{*}{OpenCV-PROSAC~\cite{chum2005matching}} & Standard &  $66.34$ & $12.53$ & $\phantom{1}3.26$ \\
   & $\nabla$-trained & $\textbf{67.22}$ & $\textbf{10.76}$ & $\phantom{1}\textbf{2.99}$ \\ \hline
    \multirow{2}{*}{MAGSAC-PROSAC~\cite{barath2020magsac++}} &Standard &  $69.09$ & $13.06$ & $163.51$ \\
    &$\nabla$-trained & $\textbf{69.94}$ & $\textbf{10.31}$ & $\textbf{128.97}$\\
    \bottomrule\\
    \end{tabular}}
    \caption{
    LoFTR performance before (standard) and after ($\nabla$-trained) trained with $\nabla$-RANSAC. Evaluating \textbf{F} matrix estimation uses three different inference methods on 12K image pairs from PhotoTourism. $\nabla$-RANSAC improves LoFTR 
    predicting accurate tentative matches and reliable confidence. 
    }
    \label{tab:loftr}
\end{table}

\section{Conclusion}
\label{sec:conclusion}

In this paper, we propose the differentiable $\nabla$-RANSAC, \ie, the first trainable randomized robust estimator with every component differentiable. 
It predicts the inlier probabilities of the input data points, exploits the predictions in a guided sampler, and estimates the model (\eg, fundamental matrix) and its quality while propagating the gradients through the entire procedure.    
$\nabla$-RANSAC leads to the most accurate epipolar geometries compared to state-of-the-art robust estimators on real-world datasets from outdoors and indoors.
Also, it is one of the fastest algorithms. 
To demonstrate its potential in unlocking the training of geometric pipelines, we train $\nabla$-RANSAC together with a recent detector-free feature matcher, LoFTR \cite{Sun_2021_CVPR}, with which we achieve improved confidence prediction and accurate robust estimation. 
We believe that the community will benefit from the algorithm.
The code and trained models are available at \url{https://github.com/weitong8591/differentiable\_ransac}.

\section*{Acknowledgements}
This research was supported
by Research Center for Informatics (project CZ.02.1.01/0.0/0.0/16\_019/0000765 funded by OP VVV),
by the Czech Technical University in Prague grant No. SGS23/173/OHK3/3T/13,
by Centers for BMK, BMAW, and the State of Upper Austria within the SCCH competence center INTEGRATE (grant no. 892418) part of the FFG COMET Competence Excellent Technologies Programme and
by the ETH Postdoc Fellowship.

\clearpage
{\small
\bibliographystyle{ieee_fullname}
\bibliography{egbib, PL}
}

\end{document}